\documentclass[10pt,twocolumn,letterpaper]{article}

\usepackage{threeparttable}
\usepackage{multirow}
\usepackage{iccv}              

\definecolor{iccvblue}{rgb}{0.21,0.49,0.74}
\usepackage[pagebackref,breaklinks,colorlinks,allcolors=iccvblue]{hyperref}

\def\paperID{*****} 
\def\confName{ICCV}
\def\confYear{2025}

\title{FusionEnsemble-Net: An Attention-Based Ensemble of Spatiotemporal Networks for Multimodal Sign Language Recognition}

\author{
Md. Milon Islam\textsuperscript{1 *},     Md Rezwanul Haque\textsuperscript{1 *},
S M Taslim Uddin Raju\textsuperscript{1},
Fakhri Karray\textsuperscript{1,2} \\
\textsuperscript{1}Department of Electrical and Computer Engineering, University of Waterloo\\
\textsuperscript{2}Department of Machine Learning, Mohamed bin Zayed University of
Artificial Intelligence \\
{\tt\small \textsuperscript{1}\{milonislam, rezwan, smturaju, karray\}@uwaterloo.ca, \textsuperscript{2}fakhri.karray@mbzuai.ac.ae}
}

\begin{document}
\maketitle

\begingroup
\renewcommand\thefootnote{}
\vspace{-1em} 

\footnotetext{
\textsuperscript{*}Equal contribution. 
Correspondence to: Md. Milon Islam \texttt{<milonislam@uwaterloo.ca>}.}

\footnotetext{
\textsuperscript{©}\textit{Proceedings of the IEEE/CVF International Conference on Computer Vision (ICCV), Honolulu, Hawaii, USA. 1st MSLR Workshop 2025. Copyright 2025 by the author(s).}}

\endgroup

\begin{abstract}
Accurate recognition of sign language in healthcare communication poses a significant challenge, requiring frameworks that can accurately interpret complex multimodal gestures. To deal with this, we propose FusionEnsemble-Net, a novel attention-based ensemble of spatiotemporal networks that dynamically fuses visual and motion data to enhance recognition accuracy. The proposed approach processes RGB video and range Doppler map radar modalities synchronously through four different spatiotemporal networks. For each network, features from both modalities are continuously fused using an attention-based fusion module before being fed into an ensemble of classifiers. Finally, the outputs of these four different fused channels are combined in an ensemble classification head, thereby enhancing the model's robustness. Experiments demonstrate that FusionEnsemble-Net outperforms state-of-the-art approaches with a test accuracy of 99.44\% on the large-scale MultiMeDaLIS dataset for Italian Sign Language. Our findings indicate that an ensemble of diverse spatiotemporal networks, unified by attention-based fusion, yields a robust and accurate framework for complex, multimodal isolated gesture recognition tasks. The source code is available at: \url{https://github.com/rezwanh001/Multimodal-Isolated-Italian-Sign-Language-Recognition}.
\end{abstract}    
\section{Introduction}
\label{sec:intro}

Sign Languages (SLs) are fully developed visual-gestural communication systems that serve as primary languages for deaf communities worldwide \cite{mineo2024sign}. In Italy, Italian Sign Language (LIS) was officially recognized in May 2021, which was an essential step in acknowledging its linguistic and socio-cultural significance \cite{fontana2021italian, caligiore-etal-2024-multisource}. Unlike spoken languages, SLs are multimodal and multilinear, incorporating simultaneous manual (e.g., handshape, movement) and non-manual (e.g., facial expression, torso posture, eye gaze) components to generate meaning \cite{chen2017adversarial, scelzi2010componenti}. As a result, automatic Sign Language Recognition (SLR) has gained increasing attention as a tool for reducing these communication gaps \cite{volterra2022italian, tornay2020hmm}. Despite this progress, many communication barriers still exist. These are especially serious in places like healthcare, where deaf patients often struggle to obtain clear and timely information during medical visits \cite{kulhandjian2019sign}. Sign language interpreters are ideal, but not always available, particularly in emergency or low-resource settings. Hence, automatic SLR has become an important area of research \cite{fontana2021italian, sincan2020autsl}.

However, developing reliable SLR systems remains a challenging task due to several interrelated factors. Firstly, SLs are inherently multimodal and multilinear, involving simultaneous and coordinated gestures that traditional sequential models often do not capture \cite{volterra2022italian}. In addition, many existing datasets are limited in scope, lacking sufficient diversity in signer demographics, environmental conditions, and sensor modalities \cite{joze2018ms}. This lack of variation makes it difficult for models to generalize between different real-world scenarios. Furthermore, in healthcare settings, the use of cameras to capture sign language raises significant concerns about patient privacy. Hospitals and clinics may be reluctant to adopt such systems due to regulatory restrictions and the need to protect patient confidentiality \cite{kulhandjian2019sign, lu2020sign}.

Moreover, recent advances in sign language recognition have explored multi-source and multimodal methods to overcome the limitations of single-modality systems. By integrating data from sensors such as RGB-D cameras \cite{huang2018video}, LiDAR \cite{mineo2024sign}, and millimeter-wave radar \cite{santhalingam2020expressive}, these approaches aim to capture complementary features that represent different dimensions of signing. Radar is valued for its ability to track movement without capturing identifiable visual information, making it suitable for privacy-preserving applications in smart healthcare \cite{caligiore-etal-2024-multisource, lu2020sign}. Multisensor datasets such as MultiMeDaLIS illustrate the benefits of combining multiple input sources to enhance the modeling of manual and non-manual components in sign language \cite{caligiore-etal-2024-multisource}. Furthermore, studies in sign language linguistics have emphasized the need for systems that ensure the diversity of the signing community, including varying levels of bilingualism and language acquisition pathways \cite{fontana2021italian, onofrio2014bilinguismo}. Although these developments demonstrate significant progress, a comprehensive architectural solution remains necessary. For robust sign language recognition, a framework must be capable of integrating multi-modal data modalities while concurrently modeling the crucial spatiotemporal features. 

To address the growing need for an accurate and robust SLR system, this research proposes FusionEnsemble-Net, an attention-based ensemble of diverse spatiotemporal networks for multimodal isolated sign language recognition. The major contributions of this paper are as follows.

\begin{itemize}
  \item We introduce FusionEnsemble-Net, an attention-based ensemble of diverse spatiotemporal networks for multimodal isolated sign recognition, fusing visual (RGB) and motion (RDM radar) data.
  
  \item We exploit an attention-based fusion approach that enables the model to automatically learn the relative importance of visual and motion data for each sign instance, generating an efficient and context-aware representation of features.
  
  \item We conduct extensive experiments on the large-scale MultiMeDaLIS dataset, demonstrating that FusionEnsemble-Net obtains a new State-of-the-Art (SOTA) test accuracy of 99.44\%. This significantly outperforms SOTA methods and builds a new performance benchmark for the task.
\end{itemize}

The remaining sections of this paper are organized as follows. Section \ref{sec:related works} presents an overview of relevant works in multimodal sign language recognition. Section \ref{sec:method} describes the proposed FusionEnsemble-Net architecture, including multimodal data processing, parallel feature extraction, and the attention-based fusion mechanism. Section \ref{sec: experiments} discusses the dataset used, implementation details, and provides a thorough analysis of our findings, which compares our model to SOTA approaches. Finally, section \ref{sec: conclusion} summarizes our findings and discusses potential directions for further research.

\section{Related Work}
\label{sec:related works}

Recent advances in SLR have used multimodal datasets and deep learning to tackle the challenges of visual-manual languages in healthcare contexts \cite{mineo2025text}. Multiple studies have explored the integration of diverse sensor data and neural models to enhance recognition accuracy and real-world applicability.

Mineo et al. \cite{mineo2024sign} introduced MultiMeDaLIS, a multimodal dataset developed for LIS recognition in medical scenarios. The dataset captured over 25,000 sign instances using synchronized data from mm-wave radar, LiDAR, RGB, RGB-D, and stereoscopic cameras to record facial expressions, hand gestures, and body movements. A structured recording protocol ensured consistency, and a custom foot-pedal system allowed participants to sign naturally. Native LIS signers reviewed the dataset and confirmed that it was accurate and culturally appropriate.  Preliminary experiments using deep learning models on individual modalities showed promising results, building a solid foundation for future AI-based sign language translation in healthcare. Caligiore et al. \cite{caligiore-etal-2024-multisource} evaluated LIS recognition in medical environments using the MultiMedaLIS dataset. The study focused on integrating RGB, depth, optical flow, and skeletal data to assess two deep learning models: Skeleton-Based Graph Convolutional Network (SL-GCN) and Spatiotemporal Separable Convolutional Network (SSTCN). The SL-GCN achieved the highest Top-1 and Top-5 accuracies of 97.98\% and 99.94\%, while SSTCN obtained 96.33\% accuracy. Although radar data was collected, it was excluded from evaluation. The results showed that multi-source input improved recognition and confirmed the suitability of the dataset for clinical SLR applications.

In another study, Fontana and Caligiore \cite{fontana2021italian} analyzed the application of natural language processing to LIS. The authors highlighted challenges related to representing visual-manual features and the absence of a standard writing system for LIS. Deaf participants played an essential role in creating more effective datasets. The study mentioned difficulties due to limited annotated resources and the complexity of non-manual signals. The authors suggested that inclusive and technology-driven methods were required to advance LIS recognition and translation. In addition, Mineo et al. \cite{mineo2025radar} presented a radar-based imaging framework for Italian Sign Language recognition, designed for medical communication. The method utilized a 60 GHz mm-wave radar sensor to capture hand motion, thereby preserving privacy by avoiding the collection of video data. The approach combined a residual autoencoder to compress radar features and a transformer classifier to recognize signs. The system was trained and tested on a new dataset of 126 LIS signs, including medical terms and alphabet letters. The proposed method achieved 93.6\% accuracy and outperformed existing radar and RGB-based approaches, showing potential for privacy-preserving healthcare applications. 

Furthermore, Vahdani et al. \cite{vahdani2024multi} proposed a multi-stream 3D Convolutional Neural Network (CNN) framework for real-time American Sign Language (ASL) recognition. The method fused multimodal features from hand gestures, facial expressions, and body poses across RGB, depth, motion, and skeleton channels. A temporal augmentation technique generated “proxy videos” to capture temporal dynamics, even with limited data. The ASL-100-RGBD dataset, which contains 100 annotated ASL signs from 22 native signers, was used to perform the experiments. The framework achieved 92.88\% accuracy on ASL-100-RGBD and demonstrated results on the ChaLearn IsoGD benchmark, thereby ensuring the effectiveness of multimodal, multichannel fusion for ASL recognition. Neverova et al. \cite{neverova2015moddrop} introduced a multi-scale, multi-modal deep learning framework for gesture detection and localization. The architecture fused intensity and depth video, containing pose and audio on three temporal scales. The ModDrop training technique enabled robust cross-modality fusion by randomly dropping channels during training, making the model resistant to noisy data. Experiments on the ChaLearn 2014 LAP dataset showed top performance, with further gains when audio was included. The model achieved a Jaccard index of 0.87, setting a new benchmark for multimodal gesture recognition.

\begin{figure*}[t]

\centerline{\includegraphics[trim={1cm 1.5cm 1.5cm 1.9cm}, scale=.52, angle=0]{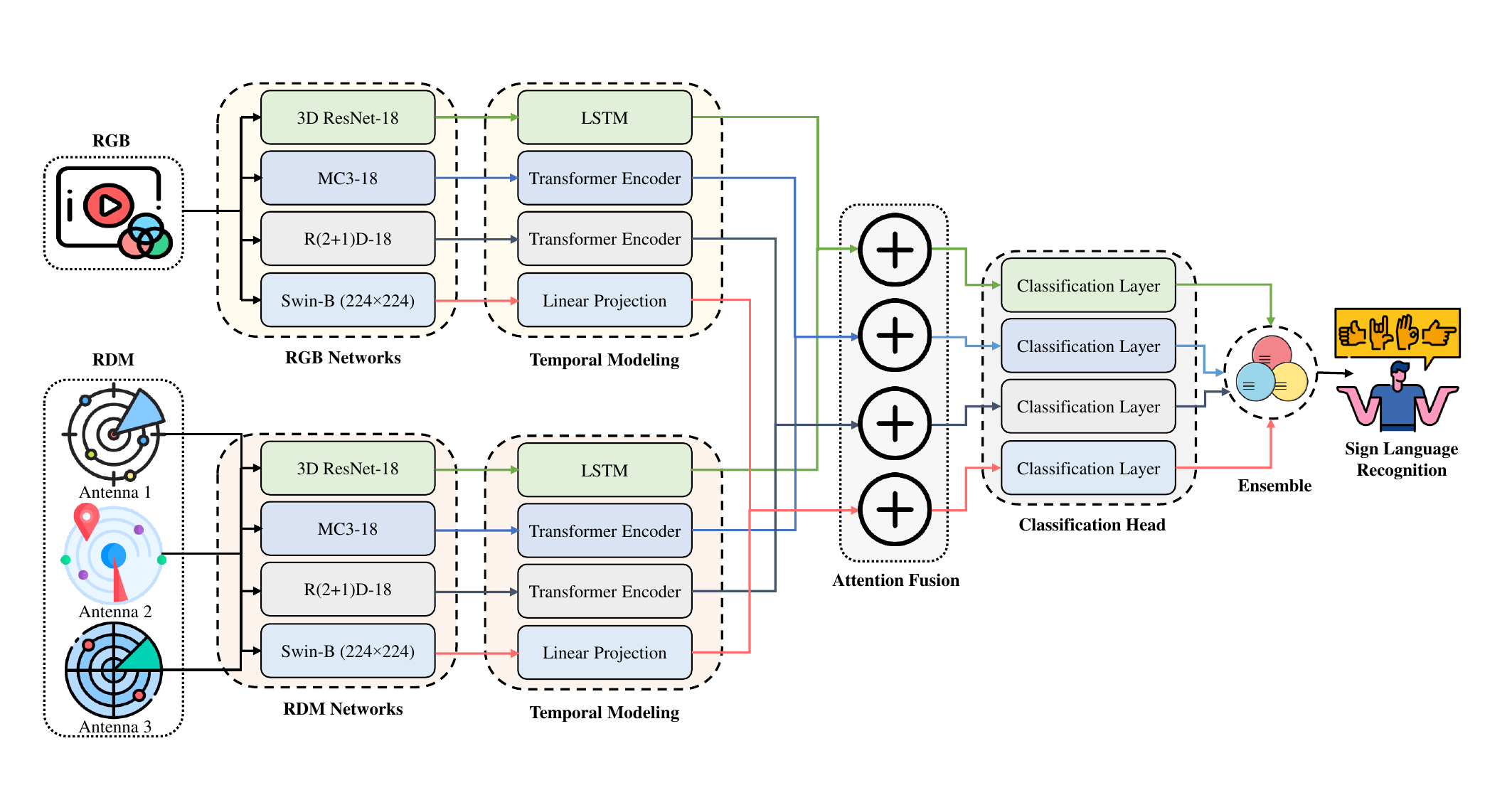}} 

\caption{Our proposed hierarchical ensemble architecture for multimodal isolated sign language recognition. The model comprises three major stages:  \textbf{Multimodal Feature Extraction}: Parallel spatiotemporal networks retrieve high-level features by processing RGB and RDM data with four different networks and temporal modeling layers. \textbf{Attention Fusion}: Feature representations are dynamically combined via an attention mechanism that fuses information across multiple modalities. \textbf{Ensemble Classification}: The four resulting feature vectors are input to independent classifiers, whose predictions are aggregated to provide a final sign language recognition output.}
\label{fig:Challenge 2}
\end{figure*}

\section{Method}
\label{sec:method}

In this section, we present FusionEnsemble-Net, our proposed hierarchical ensemble architecture for multimodal isolated sign language recognition, as illustrated in Fig. \ref{fig:Challenge 2}. The network is designed to utilize two data modalities: RGB video and Range-Doppler Map (RDM) radar data. Our method involves processing each modality through a parallel set of four different spatiotemporal networks: 3D ResNet-18, MC3-18, R(2+1)D-18, and Swin-B, to extract high-level visual and motion features. The outputs of these networks are then fed into temporal modeling layers, such as LSTMs, transformer encoders, and linear projections, which efficiently learn the dynamic sequences that characterize sign language. The attention fusion module is a key component of our model, since it effectively combines feature representations from multiple models across the two modalities. Finally, each of the four fused feature networks is sent to a distinct classification head. The final recognition is achieved by combining the results of these classifiers, which improves the overall performance of the proposed architecture. The following subsections provide a detailed description of each step in this architecture.

\subsection{Multimodal Data and Preprocessing}
\label{sec:data_streams}

Our model processes two distinct data modalities simultaneously: RGB video and RDM radar data. The input RGB video clip, represented as $V_{RGB}$ in (\ref{eq:vrgb}),  consists of standard color video clips, which provide rich visual information about handshapes, facial expressions, and body posture. The RDM modality, denoted as $V_{RDM}$ in (\ref{eq:vrdm}), is captured from three separate radar antennas. The RDM data is robust to varying lighting conditions and provides detailed motion information by measuring the velocity and range of the signer's movements.
\begin{align}
    V_{RGB} &\in \mathbb{R}^{B \times T \times C \times H \times W} \label{eq:vrgb} \\
    V_{RDM} &\in \mathbb{R}^{B \times A \times T \times C \times H \times W} \label{eq:vrdm}
\end{align}
where $B$ is the batch size, $T$ is the number of frames (time), $C$ is the number of channels, $H$ and $W$ are the frame height and width, and $A=3$ is the number of radar antennas. Both modalities capture the same sign performance synchronously. Before being fed into the network, the data are resized into a uniform dimension (e.g., $224\times224$) and normalized.

\subsection{Parallel Spatiotemporal Feature Extraction}
\label{sec:feature_extraction}

We use four separate networks to make an intelligent decision that maximize architectural variation, which is an important principle for robust ensembling. We chose models that represent distinct paradigms in video understanding: a 3D CNN (3D ResNet-18), two factorized convolutional networks (MC3-18, R(2+1)D-18), and a transformer-based model (Swin-B). This diversity assures that the models learn significant features, which improves the overall performance.

In the parallel spatiotemporal feature extraction stage, we employ a set of $N=4$ different spatiotemporal networks. The networks, listed in Table \ref{tab:feature_backbones}, include architectures like 3D ResNet-18 \cite{tran2015learning}, MC3-18  \cite{tran2018closer}, R(2+1)D-18 \cite{tran2018closer}, and the transformer-based Swin-B \cite{liu2021swin}. For each modality $i \in \{1, ..., N\}$, a network function $\Phi_i$ extracts features from both modalities. For the RGB, the feature extraction is computed as in (\ref{eq:F_RGB}).
\begin{equation}
    F_{RGB,i} = \Phi_{i,RGB}(V_{RGB})
    \label{eq:F_RGB}
\end{equation}
For RDM, we process the data from each of the $A=3$ antennas through the network and then average the resulting features to obtain a unified representation as in (\ref{eq:RDM_i}).
\begin{equation}
    F_{RDM,i} = \frac{1}{A} \sum_{a=1}^{A} \Phi_{i,RDM}(V_{RDM,a})
    \label{eq:RDM_i}
\end{equation}
where $V_{RDM,a}$ is the data from the $a$-th antenna.

The extracted features, $F_{RGB}$ and $F_{RDM}$,  are passed to a temporal modeling layer, denoted by $\mathcal{T}_i$ (e.g., an LSTM/transformer encoder / linear projection), which are represented in (\ref{eq:H_RGB}) and (\ref{eq:H_RDM}), to capture sequential dependencies.
\begin{align}
    H_{RGB,i} &= \mathcal{T}_{i,RGB}(F_{RGB,i}) \label{eq:H_RGB} \\
    H_{RDM,i} &= \mathcal{T}_{i,RDM}(F_{RDM,i}) \label{eq:H_RDM}
\end{align}
This process yields modality-specific temporal feature vectors, $H_{RGB,i}$ and $H_{RDM,i}$, for each of the $N$ parallel networks. These layers are designed to learn the sequential dependencies within the signs, which is crucial for distinguishing between signs with similar but temporally distinct movements.

We employ a variety of temporal layers (LSTM, transformer, and linear projection) to enhance model diversity.  This heterogeneity enables the ensemble to capture a wider range of temporal dynamics, hence increasing overall performance.

\begin{table}[t]
\centering

\begin{tabular}{lll}
\hline
\textbf{Methods} & \textbf{RGB} & \textbf{RDM} \\
\hline
3D ResNet \cite{tran2015learning} & 3D ResNet-18 & 3D ResNet-18 \\
MC3 \cite{tran2018closer} & MC3-18 & MC3-18 \\
R(2+1)D \cite{tran2018closer} & R(2+1)D-18 & R(2+1)D-18 \\
Swin-B \cite{liu2021swin} & Swin-B (224x224) & Swin-B (224x224) \\
\hline
\end{tabular}
\caption{Specification of the feature extraction networks utilized in our proposed multimodal architecture. Each model is applied synchronously on both the visual and radar data modalities to retrieve diverse spatiotemporal features.}
\label{tab:feature_backbones}

\end{table}

\subsection{Attention-Based Feature Fusion}
\label{sec:attention_fusion}

For each stream, temporal feature vectors from RGB and RDM are concatenated along the feature dimension. A self-attention module then combines this concatenated vector, dynamically re-weighting visual and motion features to generate a single fused representation.

To fuse the information from both modalities, we first concatenate the temporal feature vectors for each modality $i$ as in (\ref{eq:H_concat}).
\begin{equation}
    H_{concat,i} = [H_{RGB,i}; H_{RDM,i}]
    \label{eq:H_concat}
\end{equation}
where $[;]$ denotes concatenation. These concatenated features are then fed into an attention fusion module, $\Psi_i$. We utilize a self-attention mechanism, which enables the model to dynamically weigh the importance of different feature components. The attention output is calculated using the scaled dot product attention formula based on (\ref{eq:attn_qkv}).
\begin{equation}
    \text{Attention}(Q, K, V) = \text{softmax}\left(\frac{QK^T}{\sqrt{d_k}}\right)V
    \label{eq:attn_qkv}
\end{equation}
where the Query ($Q$), Key ($K$), and Value ($V$) matrices are linear projections of input $H_{concat,i}$, and $d_k$ is the dimension of the key vectors. This fusion generates a feature vector $F_{fused,i}$ for each network derived in (\ref{eq:F_fused}).
\begin{equation}
    F_{fused,i} = \Psi_i(H_{concat,i})
    \label{eq:F_fused}
\end{equation}
The attention-based fusion mechanism dynamically weighs the features of each modality, allowing the model to decide which modality contains more relevant information for a particular sign.

\subsection{Ensemble Classification Head}
\label{sec:ensemble_classification}

In the final stage, each of the fused feature vectors $N=4$ $F_{fused,i}$ is passed to an independent classification head $C_i$. Each head generates a vector of logits, $L_i$ as per (\ref{eq:L_i}), for the sign classes.
\begin{equation}
     L_i = F_{fused,i} W_{C,i}^T + b_{C,i}
    \label{eq:L_i}
\end{equation}
where $W_{C,i}$ is the weight matrix and $b_{C,i}$ is the bias vector of the linear classifier for the $i$-th network. 
The logits are then converted into probability distributions using the softmax function, $\sigma$ as calculated in (\ref{P_i}).
\begin{equation}
    P_i = \sigma(L_i)
    \label{P_i}
\end{equation}
During training, these logits $L_i$ are used directly to compute the cross-entropy loss $\mathcal{L}_{CE}$ against the ground-truth labels $y_{true}$, derived in (\ref{eq:loss}).
\begin{equation}
    \text{Loss} = \sum_{i=1}^{N} \mathcal{L}_{CE}(L_i, y_{true})
    \label{eq:loss}
\end{equation}
The entire architecture has been trained end-to-end.  The total loss, aggregated across all heads, is backpropagated to optimize all components at the same time, enabling parallel streams to learn significant features. The final prediction of FusionEnsemble-Net is obtained by averaging the probabilities from all $M$ classifiers according to (\ref{eq:P_final}). 
\begin{equation}
    P_{final} = \frac{1}{M} \sum_{i=1}^{M} P_i
    \label{eq:P_final}
\end{equation}
The predicted class, $\hat{y}$ as per (\ref{eq:y_hat}), is the one with the highest final probability.
\begin{equation}
    \hat{y} = \underset{k}{\operatorname{argmax}}(P_{final,k}) 
    \label{eq:y_hat}
\end{equation}
This ensemble method makes the model more robust and less prone to errors from any single feature extractor. By combining the predictions from diverse models, we obtain a more accurate and reliable sign language recognition system.

\section{Experiments}
\label{sec: experiments}

\subsection{Dataset}

All experiments were conducted on the MultiMedaLIS dataset \cite{caligiore-etal-2024-multisource, mineo2025mslr}, which is the basis for the SignEval 2025 recognition challenge \cite{luqman2025signeval}. This is a multimodal dataset for recognizing isolated Italian Sign Language, designed for medical contexts. It contains 126 unique signs, including 100 medical-related terms and 26 letters of the LIS alphabet. A key feature of this dataset is its inclusion of multiple synchronized data sources for each sign performance, including RGB, depth, skeletal, and the RDM radar data. We follow the data splits provided with the dataset for training, validation, and testing.

\subsection{Implementation Details}
Our proposed FusionEnsemble-Net was implemented using PyTorch. The spatiotemporal networks were initialized with weights pre-trained on large-scale video or image datasets (Kinetics-400 for 3D CNNs and ImageNet for Swin-B) to leverage transferred knowledge.

The models were trained using the AdamW optimizer with a cosine annealing learning rate scheduler. We used a batch size of 4 and trained the models on two NVIDIA A6000 GPUs. The temporal modeling layers consisted of 2-layer LSTMs with a hidden layer size of 512 or transformer encoders with 8 attention heads. The final prediction was generated by averaging the output of the four classification heads. Training each model for 25 epochs required approximately 44 hours (1 hour 45 minutes per epoch).

\subsection{Evaluation Metrics}
We evaluated the performance of our models using the standard top-1 accuracy metric. We report the accuracy on both the validation and test sets to provide a comprehensive assessment of our model's performance and its ability to generalize.

\begin{table}[b]
  \centering
  \begin{threeparttable}
  \resizebox{\columnwidth}{!}{%
  \begin{tabular}{llll}
    \hline
    \textbf{Methods} & \textbf{Modality} & \textbf{Valid} & \textbf{Test} \\
    \hline
    SL-GCN \cite{caligiore-etal-2024-multisource} & \multirow{4}{*}{RGB} & - & 97.98 \\
    SSTCN \cite{caligiore-etal-2024-multisource} &  & - & 96.33 \\
    ResNet(2+1)D Optical Flow \cite{caligiore-etal-2024-multisource} &  & - & 56.31 \\
    ResNet(2+1)D Frame \cite{caligiore-etal-2024-multisource} &  & - & 97.29 \\
    ResNet(2+1)D Encoding HHA \cite{caligiore-etal-2024-multisource} & Depth & - & 88.04 \\
    \hline
    \multirow{6}{*}{AutoTrans-RDMNet \cite{mineo2025radar}} & RDM & - & 88.3 \\
     & 3×RDM & - & 91.7 \\
     & MTI & - & 84.9 \\
     & 3×MTI & - & 86.1 \\
     & RDM+MTI & - & 91.4 \\
     & 3×RDM+3×MTI & - & 93.6 \\
    \hline
    3D ResNet & \multirow{5}{*}{RGB+3×RDM} & 96.58 & 96.58 \\
    MC3 &  & \underline{98.96} & \underline{99.06} \\
    R(2+1)D &  & 96.94 & 97.34 \\
    Swin-B &  & 94.24 & 94.42 \\
    \textbf{FusionEnsemble-Net} &  & \textbf{99.37} & \textbf{99.44} \\
    \hline
  \end{tabular}
  }
  \begin{tablenotes}
      \footnotesize
      \item[*] HHA=Height, Horizontal disparity, Angle, and MTI=Moving Target\\ Indications.
  \end{tablenotes}
  \caption{Performance comparison of our proposed FusionEnsemble-Net with SOTA approaches on MultiMedaLIS dataset. The results for each network within our multimodal system are also reported. Our proposed ensemble architecture outperforms all other methods on both the Validation (Val) and test sets. Bold values denote the best performance, while underlined values denote the second-best.}
  \label{tab:model_performance}
  \end{threeparttable}
\end{table}

\subsection{Results}
The performance of our proposed FusionEnsemble-Net and its comparison with SOTA methods are presented in Table \ref{tab:model_performance}. To compare our findings, we selected SL-GCN \cite{caligiore-etal-2024-multisource} and AutoTrans-RDMNet \cite{mineo2025radar}, which achieve high performance in the SOTA approaches on this dataset. We benchmark against SL-GCN \cite{caligiore-etal-2024-multisource}, a graph-based network that achieved 97.98\% accuracy using skeleton data. Simultaneously, we include AutoTrans-RDMNet \cite{mineo2025radar}, the radar-centric framework with 93.6\% accuracy, while considering both RDM data and moving target indications, which serves as the key privacy-sensitive baseline.

We analyze the individual performance of each of the four parallel networks within our architecture. A key observation is that the MC3 network pair reports the highest accuracy among the single baselines, achieving test and validation accuracies of 99.06\% and 98.96\%, respectively. This suggests that its hybrid architecture, which combines 2D and 3D convolutions, is suitable for extracting complementary spatiotemporal features from the visual (RGB) and motion (RDM) data. The other networks, including 3D ResNet and R(2+1)D, also demonstrate competitive performance, validating our multimodal architecture even at the individual baseline level.

It is revealed that by aggregating the predictions from all four diverse networks, the model achieves a final test accuracy of 99.44\% and a validation accuracy of 99.37\%. This outcome establishes a new SOTA on the MultiMedaLIS dataset and demonstrates the significant value of our proposed ensemble approach. The significant performance improvement over the best single network (MC3) validates our core hypothesis: that an attention-based ensemble of diverse spatiotemporal networks is highly effective for this task. The ensemble efficiently leverages this diversity, resulting in a more robust and generalizable model that is less susceptible to the biases inherent in any single architecture.

\section{Conclusion}
\label{sec: conclusion}

In this paper, we introduced FusionEnsemble-Net, a novel attention-based ensemble of diverse spatiotemporal networks designed to achieve SOTA performance in multimodal isolated sign language recognition. The proposed architecture effectively fuses visual information from RGB video with motion data from RDM radar. By processing these modalities through four diverse spatiotemporal networks and integrating their features with an attention mechanism, our model efficiently captures the complex dynamics of signing. Our key finding is that this architecture obtains a new SOTA accuracy of 99.44\% on the MultiMeDaLIS dataset, demonstrating the strength of combining multimodal inputs with a diverse set of feature extractors. The impacts of the proposed architecture are significant, indicating a clear direction for developing reliable assisted communication systems in healthcare context.

The major limitation of our model is that it was evaluated on isolated signals from signers and its high computational complexity may pose issues for real-time deployment on resource-constrained devices. The potential future work involves expanding our system to address the issue of continuous conversational sign language recognition. Lastly, we intend to investigate model compression approaches, including knowledge distillation, TinyML to develop a more lightweight and efficient version of FusionEnsemble-Net without a significant reduction in performance, bringing it closer to practical application.

{
    \small
    \bibliographystyle{ieeenat_fullname} 
    \bibliography{main}
}

\end{document}